\documentclass[11pt,a4paper]{article}
\usepackage[hyperref]{eacl2021}
\usepackage{times}
\usepackage{latexsym}

\usepackage{booktabs}
\usepackage{graphicx}
\usepackage{CJKutf8}
\usepackage[linesnumbered,ruled]{algorithm2e}
\usepackage{amsmath}
\usepackage{microtype}
\usepackage{bbm}
\aclfinalcopy 

\usepackage{algpseudocode}
\usepackage{listings}
\definecolor{codeBackground}{rgb}{1.00,1.00,1.00}

\lstnewenvironment{python}[2]
{
  \lstset{
    title={#1},
    backgroundcolor=\color{codeBackground},
    numbers=left,
    stepnumber=1,
    basicstyle=\ttfamily\footnotesize,
    keywordstyle=\color{blue}\bfseries,
    showstringspaces=false,
    xleftmargin=10pt,
    language=python,
    frame=t,
  }
}
{}
\newcommand{\glm}{GLMask}

\title{Gradient-guided Loss Masking for Neural Machine Translation}

\author{
\textbf{Xinyi Wang}$^{1}$,
\textbf{Ankur Bapna}$^{2}$,
\textbf{Melvin Johnson}$^{2}$,
\textbf{Orhan Firat}$^{2}$
\\
$^{1}$Language Technology Institute, Carnegie Mellon University \\
$^{2}$Google Research \\
\texttt{xinyiw1@cs.cmu.edu,\{ankurbpn, melvinp, orhanf\}@google.com}
}

\date{}

\begin{document}
\maketitle
\begin{abstract}
    To mitigate the negative effect of low quality training data on the performance of neural machine translation models, most existing strategies focus on filtering out harmful data before training starts. In this paper, we explore strategies that dynamically optimize data usage during the training process using the model's gradients on a small set of clean data. At each training step, our algorithm calculates the gradient alignment between the training data and the clean data to mask out data with negative alignment. Our method has a natural intuition: good training data should update the model parameters in a similar direction as the clean data. Experiments on three WMT language pairs show that our method brings significant improvement over strong baselines, and the improvements are generalizable across test data from different domains.    
\end{abstract}
\section{Introduction\label{sec:intro}}
The quality of training data has a large effect on the performance of neural machine translation~(NMT) models~\citep{noise_impact_nmt}. Parallel sentences crawled from the web and aligned from multilingual documents~\cite{para_crawl} provide a large supply of training data to boost the performance of NMT, but automatically extracted training examples often have noise that could hurt the model performance. Designing a good strategy to utilize the training data with varying quality levels is essential to improving the performance of NMT models. 

Prior methods mainly focus on filtering data before training starts~\citep{select_lm_moore_lewis, dynamic_data_selection,wang-etal-2018-denoising}. \citet{dual_ce_filter} proposed selecting clean training data by using a heuristic threshold on the perplexity difference between the forward and backward translation models. \citet{wang-etal-2018-denoising} propose a curriculum-based data selection strategy using a small set of clean trusted data. 


In this paper, we examine the problem of optimizing the training data usage from a different angle. Instead of filtering data before the training, we utilize the gradient information of the data \textit{during} the training process to guide the data usage. The intuition behind our method is natural: given a very small set of clean or trusted data~(which is generally easy to obtain in practice), the preferred training data should have similar gradient direction as the gradient of the clean data, so that updating the model parameters on this training data would improve the model performance on the clean data. Based on this intuition, we propose gradient-guided loss masking~(\glm), which calculates the gradient alignment between each training example and the clean data, and simply masks out the loss of the training examples that have negative gradient alignments. \glm~is inspired by methods that optimize training data usage using meta-learning. \citet{learn_to_weight}~calculates a weighting of noisy training data that minimizes the model loss on clean data for image classification tasks. \citet{multi_dds,dds}~optimizes the training data distribution of multilingual data such that the model loss on the development set is minimized. Our method, on the other hand, uses a simple gradient alignment signal to determine the parallel sentence pairs or target words to mask out during NMT training.   

We test \glm~on the standard WMT English-German, English-Chinese and English-French translation tasks, using WMT test sets from prior years as the small clean data. \glm~brings significant improvements over strong baselines on all three language pairs. We further evaluate the trained models on the IWSLT test sets which are sampled from a different domain, and demonstrate that \glm~also delivers improvements on these out of domain sets. This shows that improvements from \glm~generalize to data from different domains, beyond the domain of the clean data.

\section{Method\label{sec:method}}
To explain our method, we first provide a mathematical framework for the problem it aims to address. 
\subsection{Problem with training data} 
To train a NMT model with parameters $\theta$ that translates from a source language $S$ to a target language $T$, we want to find the optimal model parameters $\theta^*$, which minimizes the loss function $\ell(x,y, \theta)$ over the \textit{true} distribution of the parallel data from $S$-$T$, denoted by $P(S, T)$. However, in practice, one usually only has access to a limited number of parallel training sentences sampled from the training distribution $P_\text{train}(x, y)$. The standard training approach finds the $\theta^*$ that minimizes the loss function $\ell(\cdot; \theta)$ over this training distribution:
\begin{align}
\label{eqn:train_obj_orig}
    J_{\text{train}}(\theta) = \mathbf{E}_{x, y \sim P_\text{train}(x, y)} [\ell(x, y;\theta)]
\end{align}

Problems arise when the training distribution $P_\text{train}(x, y)$, from which we draw training sentences, has discrepancies with $P(S, T)$, the true distribution of the parallel data from $S$-$T$. For example, some samples from the training data might be noisy and might be detrimental to final model performance on the clean data~\citep{noise_impact_nmt}.

To remedy this training data discrepancy, one strategy is to collect a small set of high quality data drawn from distribution $P_\text{clean}(x, y)$, which is closer to the true data distribution $P(S, T)$, to guide training on the large noisy training data. With the help of this clean data, we want to train the model using data sampled from $P_\text{train}(x, y)$, while the loss over the clean data is minimized:  
\begin{align}
    J_\text{clean}(\theta) = \mathbf{E}_{x, y \sim P_\text{clean}(x, y)} [\ell(x, y;\theta)]
\end{align}

\subsection{Gradient-guided loss masking}
We propose gradient-guided loss masking~(\glm), which uses gradient information to mask out the training examples that can be harmful for minimizing the model's loss on the clean data. \glm~modifies the training objective $J_\text{train}(\theta)$ such that optimizing the model parameter over this objective also optimizes the objective over the small clean data $J_\text{clean}(\theta)$. The intuition behind our approach is that a training example $x, y \sim P_\text{train}(x, y)$ is more likely to minimize the loss over the clean data, if the gradient of this training data is in the same direction as the gradient of the clean data. Formally, we use the dot product between these two gradients as their alignment
\begin{align}
    g(x, y) = \nabla_{\theta} \ell(x, y; \theta)^\top \cdot \nabla_\theta J_\text{clean}(\theta)
\end{align}
A negative $g(x, y)$ indicates that the training example has a conflicting gradient with the clean data, which might be detrimental for the model's performance on the clean data. To optimize the model objective on the clean data, we can mask out the examples that have negative $g(x, y)$ with the clean data. That is, we modify the training objective in \autoref{eqn:train_obj_orig} to 
\begin{align}
   J_\text{train}(\theta) = \mathbf{E}_{x, y \sim P_\text{train}(x, y)}\left[m(x, y) \cdot \ell(x, y) \right]
\end{align}
where $m(x, y) = \mathbbm{1}[g(x, y) > 0]$. The masking term assigns a weight of 0 to examples that have negative gradient alignment with the clean data.

\paragraph{Implementation} The pseudo code for \glm~is in \autoref{alg:alg}. Note that calculating the loss mask at \autoref{alg:grad_prod} requires calculating the gradient alignment of each training example and the gradient of the clean data, which is potentially very expensive. We use a technique that allows us to efficiently compute this value with only two additional backward passes, which is supported by modern deep learning libraries such as Tensorflow~\cite{abadi2016tensorflow}\footnote{We use the technique introduced here: \url{https://j-towns.github.io/2017/06/12/A-new-trick.html}. The function in Tensorflow is provided in \autoref{sec:code_tf}}. In our preliminary experiments, using \glm~at the later stage of training works as well as or even better than using it from the beginning. Therefore, we use \glm~for the last $20\%$ of training steps, which further decreases the overall training overhead of our method.

\begin{algorithm}[t!]
\small
\SetAlgoLined
\DontPrintSemicolon
\SetKwInOut{Input}{Input}
\SetKwInOut{Output}{Output}
\SetCommentSty{itshape}
\SetKwComment{Comment}{$\triangleright$ }{}
\Input{Training corpus $\mathcal{D}_{\text{train}}$; clean data $\mathcal{D}_{\text{clean}}$ 
}
\Output{The converged model $\theta^*$}
   
  \While{not converged}{
    
    \Comment{Sample a batch of training data}
    $(x_1, y_1)...(x_B, y_B) \sim \mathcal{D}_\text{train}$
    
    \Comment{Sample a batch of clean data}
    $(x'_1, y'_1)...(x'_B, y'_B) \sim \mathcal{D}_\text{clean}$
    
    \Comment{Calculate data mask}
    $g' \leftarrow \nabla_{\theta_t} \frac{1}{B} \sum_{i=1}^{B'} \ell(x'_i, y'_i; \theta_t)$
    
    $m(x_i, y_i)$ \\
    $\leftarrow \mathbbm{1}(g'^\top \nabla_\theta \ell(x_i, y_i; \theta_t)>0) \text{  for i in 1...B} \label{alg:grad_prod}$
    
    \Comment{Calculate masked objective}
    $g_\text{train} \leftarrow \nabla_{\theta_t} \frac{1}{B} \sum_{i=1}^{B} m(x_i, y_i) \ell(x_i, y_i; \theta_t)$
    
    $\theta_{t+1} \leftarrow \text{Update}(\theta_t, g_\text{train})$
    
  }
\caption{ Training with \glm}
\label{alg:alg}
\end{algorithm}
\section{Experiment \label{sec:exp}}
\subsection{Dataset and setup}
\paragraph{Data} We use parallel data from the WMT evaluation campaign. To verify the effectiveness of our approach, we test it on three language pairs with varying amounts of resources: English to German~(en-de), English to Chinese~(en-zh), and English to French~(en-fr). 

The en-de experiments are conducted using the WMT'14 training data with about 4 million parallel sentences. We use newstest2013 as the validation set, and newstest2014 as the test set. The en-zh experiments use the WMT'17 training data with around 22 million parallel sentences. We use newsdev2017 as the validation set and newstest2017 as the test set. The en-fr experiments use the WMT'14 training data with about 40 million parallel sentences. We use newstest2013 as validation set and newstest2014 as the test set. We use sacreBLEU~\cite{sacrebleu} to evaluate all our models.

\paragraph{Model and preprocessing} We use the standard Transformer base model~\cite{transformer}, with 6 layers and 8 attention heads. The dropout rate is set to 0.1 and we use label smoothing of 0.1. For all datasets, we process the data using sentencepice~\cite{sentencepiece} with a vocabulary size of 40k. Parallel sentences with length longer than 200 word pieces are filtered out during training.

\paragraph{Construction of the clean data}
\glm~requires a small set of clean data to guide model training. Here we simply use previous years' WMT evaluation sets as our clean data. For en-de and en-fr, we concatenate the news test sets from 2010 to 2012 as the clean data, which contains approximately 8k sentences. For en-zh, we re-use our validation set, newsdev2017, as the clean data, which has about 2k sentences. In practice, it is generally reasonable to obtain a small amount of high quality annotated data.

\subsection{Baselines}
We compare with two baselines: 1) vanilla: standard transformer model trained on all parallel data; 2) finetune: we finetune the vanilla model on the small cleaned dev set used for calculating loss masking in our method.

For \glm, we use two variations of loss masking: 1) sentence level~(\glm-sent): we mask out the loss of the parallel sentences in a batch; 2) word level~(\glm-word): we mask out the loss of each individual subwords on the target side. 

\subsection{Results}
We evaluate the baselines and our methods on the WMT test sets, and document the results in \autoref{tab:wmt_results}. First, we notice that finetuning is a strong baseline,  especially when the training data is relatively diverse. It does not lead to improvement for en-de, which has the least amount of training data, but it improves over the vanilla model by 1 BLEU for the higher resourced en-zh and en-fr language pairs. \glm~improves over the strong finetune baseline for all three language pairs. Specifically, using \glm~on the word level consistently outperforms masking on the sentence level. 
\begin{table}[]
    \centering
    \begin{tabular}{l|ccc}
    \toprule \
         &  en-de & en-zh & en-fr \\
    \midrule \
     Vanilla  & 27.29 & 32.99 & 39.30 \\
     Finetune  & 27.26 & 33.99 & 40.33 \\
     \midrule \
     \glm-sent & 27.49 & 34.19 & 40.34 \\
     \glm-word & \textbf{27.94} & \textbf{34.96} & \textbf{40.66} \\
    \bottomrule 
    \end{tabular}
    \caption{Results on the WMT test set.}
    \label{tab:wmt_results}
\end{table}

\begin{table}[]
    \centering
    \begin{tabular}{l|ccc}
    \toprule \
         &  en-de & en-zh & en-fr \\
    \midrule \
     Vanilla  & 28.79 & 25.64 & 42.24 \\
     Finetune  & 28.54 & 25.17 & 42.02 \\
     \midrule \
     \glm-sent & 28.74 & 25.60 & \textbf{42.74} \\
     \glm-word & \textbf{29.22} & \textbf{26.10} & 42.48 \\
    \bottomrule 
    \end{tabular}
    \caption{Results on the IWSLT test set.}
    \label{tab:iwslt_results}
\end{table}
\paragraph{Out-of-domain generalization results}
\glm~utilizes a small clean dataset to guide the training of NMT models. One potential criticism of this design choice is that the model might overfit to this small data. To examine how well the model trained using \glm~generalizes to new domains, we construct an additional test set using the IWSLT~\citep{iwslt} data from TED talks. Since the small clean data we use is drawn from the news domain, the model performance on the IWSLT test sets is a good indicator how well \glm~generalizes to out-of-domain data. We aggregate the test sets from IWSLT\footnote{\url{https://wit3.fbk.eu/}} 2011 to 2015 to construct a large test set with about 5k sentence pairs.    

We evaluate the performance of all models on the IWSLT test set and record the results in \autoref{tab:iwslt_results}. Although finetuning outperforms the vanilla model on the WMT test sets from the news domain, it does not improve on the IWSLT test sets. This indicates that finetuning on the small clean data might overfit to the specific domain of the clean data. On the other hand, \glm~ improves over the vanilla model on the out-of-domain test set for all three language pairs without any supervision for the IWSLT domain.

\subsection{What data gets masked out?}
\begin{figure}
    \centering
\resizebox{0.3\textwidth}{!}{
    \includegraphics[width=0.4\textwidth]{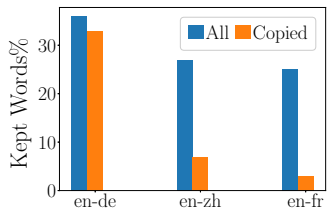}
    }
    \caption{Ave. percentage of words that are not masked out for all examples, and copied examples. Examples where source and target sentences are identical are masked out more.}
    \label{fig:copied_sentences}
\end{figure}

\begin{figure}
    \centering
\resizebox{0.3\textwidth}{!}{
    \includegraphics[width=0.4\textwidth]{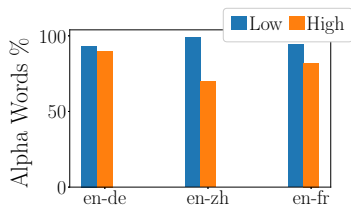}}
    \caption{Percentage of alphabetical words for the lowest and highest masked out words. Words that are masked out more tend to have less semantically rich alphabetical words.}
    \label{fig:alpha_words}
\end{figure}

In this section, we analyze the training sentences and words that tend to be masked out by \glm. We use a converged vanilla model to calculate the gradient alignments of a subset of training data, with about 250k sentence pairs. 

First, we examine the sentence pairs using the percentage of words that are not masked for each sentence. In \autoref{fig:copied_sentences}, we plot the average percentage of unmasked words in all the sampled training examples~(All), and examples that have identical source and target sentences~(Copied). En-de has higher percentage of unmasked, probably because it has a relatively cleaner training set with the least amount of data. For all three language pairs, especially the more noisy en-zh and en-fr, copied examples have much less unmasked words than average. This indicates that \glm~is able to filter out copied sentences, which is known to be one of the most harmful categories of noise for NMT~\cite{noise_impact_nmt}.   

Next, we analyze the type of target words that tend to be masked out by \glm. We sort the target words by the percentage of being masked, and compare the 1k words with the highest masking rate and the 1k words with the lowest masking rate. Alphabetical words, or words with only alphabetical characters, usually have richer semantic meaning than words that contain numbers or symbols. In \autoref{fig:alpha_words}, we compare the percentage of alphabetical words for the two group of words with the highest and lowest masking rates.  For all three languages, the words that have a higher masking rate tend to have less alphabetical words, which indicates that the words having clearer and richer meanings are more favored by \glm.   

We show some training examples and the target words that are masked out by \glm~in \autoref{tab:trans_examples}. Our method masks out the punctuation words in the target that do not align with the source sentences.

\begin{table}[t]
\centering
\resizebox{0.5\textwidth}{!}{
\begin{tabular}{c|l|l}
\toprule
& \textbf{Src} & \textbf{Trg} \\
\midrule
\textbf{en-de} & Neither have you , I hope .
& Sie \textcolor{red}{doch} hoffentlich auch nicht \textcolor{red}{?}   \\
\midrule
\textbf{en-zh} & What about you  r &
\begin{CJK*}{UTF8}{gbsn} 你认为  \textcolor{red}{呢?  r} \end{CJK*} \\
\bottomrule
  \end{tabular}}
  \caption{Examples training data. Red words are masked out.}
  \label{tab:trans_examples}
\end{table}

\section{Conclusion\label{sec:conclusion}}
In this paper, we evaluate a strategy to dynamically mask out unhelpful data during NMT training . We propose \glm, a simple method that uses the gradient alignments between the training data and a small clean dataset to improve data usage. Experiments show that our method not only brings significant improvements on three WMT datasets, but also improves out-of-domain performance.

\bibliography{main}
\bibliographystyle{acl_natbib}

\appendix
\onecolumn{
\section{Appendix}
\subsection{\label{sec:code_tf}Source Code for Training with GLMask in TensorFlow}
\begin{python}{\texttt{Get masked training gradient in TensorFlow}}

import tensorflow as tf
def get_train_gradient(train_loss, valid_loss, model):
  """Sample a batch of corrupted examples from sents.

  Args:
    train_loss: Tensor [batch_size, n_steps]. The loss for
    each training example in a batch.
    valid_loss: Tensor [1]. The aggregated loss for a batch of valid data. 
    model: the parameters of the NMT model.

  Returns:
    train_grad: Tensor [batch_size, n_steps]. The training gradient after masking.
  """
  # A dummy variable to assist calculating gradient dot product
  z = tf.ones(train_loss.shape)
  
  train_grad = tf.gradient(train_loss, model, grad_ys=z)
  valid_grad = tf.gradient(valid_loss, model)
  # dot product between train_grad and vali_grad
  dot_prod = tf.gradient(train_grad, z, grad_ys=valid_grad)
  
  # mask out the examples with negative gradient alignment
  gradient_mask = tf.greater(dot_prod, tf.zeros(dot_prod.shape))
  
  train_grad = train_grad * gradient_mask
  return train_grad
\end{python}
}
\end{document}